\documentclass[10pt,twocolumn,letterpaper]{article}

\usepackage[pagenumbers]{iccv}          
\usepackage{xcolor}
\usepackage{multirow}
\usepackage{bbding}

\definecolor{iccvblue}{rgb}{0.21,0.49,0.74}
\usepackage[pagebackref,breaklinks,colorlinks,allcolors=iccvblue]{hyperref}

\def\paperID{10820} 
\def\confName{ICCV}
\def\confYear{2025}

\title{LangScene-X: Reconstruct Generalizable 3D Language-Embedded Scenes \\ with TriMap Video Diffusion}

\author{Fangfu Liu$^{1}$, Hao Li$^{2}$, Jiawei Chi$^{1}$, Hanyang Wang$^{1}$, Minghui Yang$^{3}$, Fudong Wang$^{3}$, Yueqi Duan$^{1}$\footnotemark[2]\\
$^{1}$Tsinghua University, 
$^{2}$NTU, $^{3}$Ant Group}

\begin{document}
\twocolumn[{
 \renewcommand\twocolumn[1][]{#1}
\maketitle
 \thispagestyle{empty}
 \pagestyle{empty}
 \begin{center}
     \captionsetup{type=figure}
    \vspace{-2em} \includegraphics[width=1\linewidth]{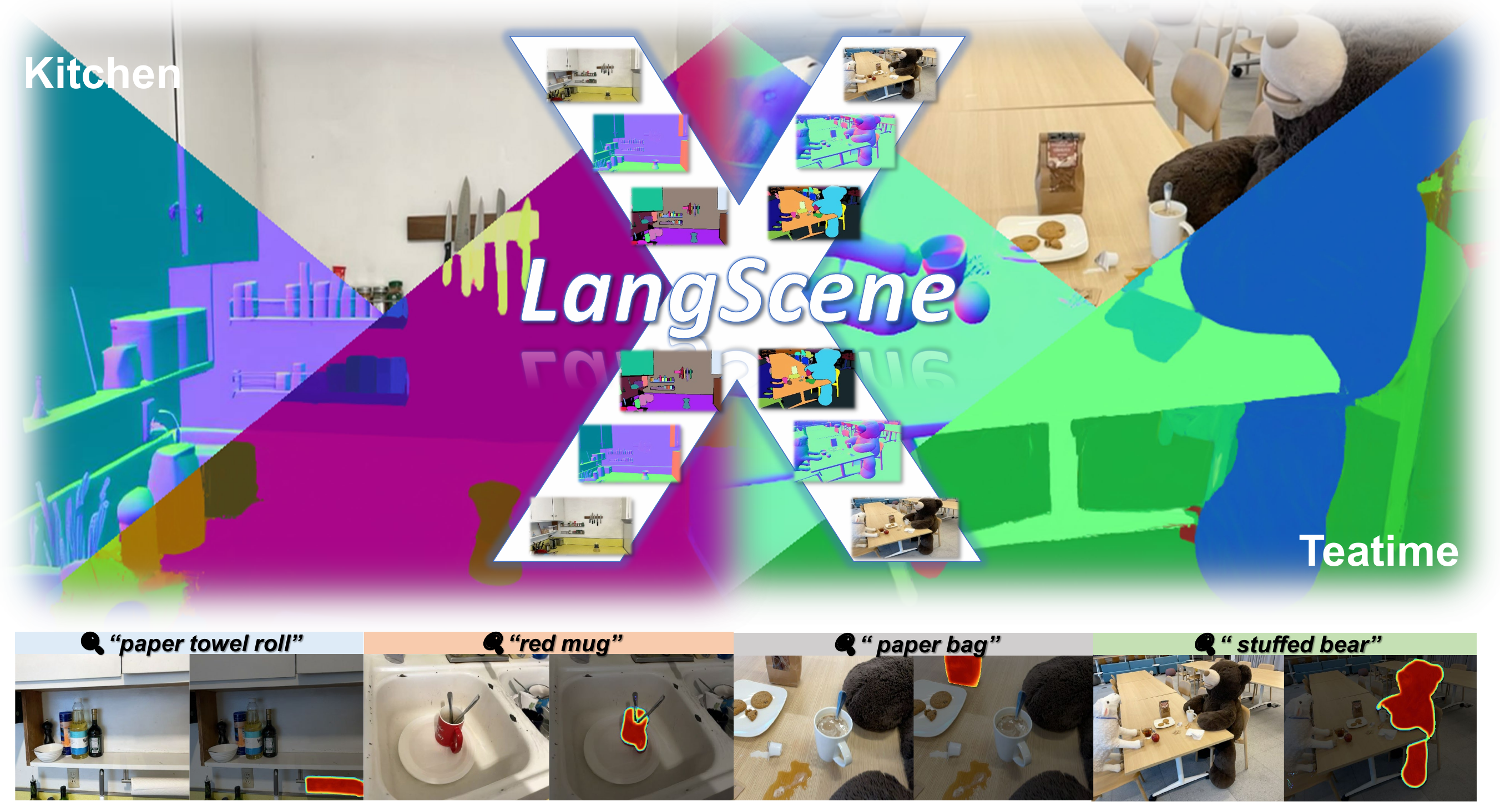}
    \vspace{-2em}
     \captionof{figure}{\textbf{LangScene-X}: Given sparse views as input (\eg, as few as two images), we design a generative paradigm to build the 3D generalizable language-embedded surface fields with TriMap video diffusion and language quantized compressor (LQC), which supports open-ended language queries in any 3D scenes. For example, given a prompt like ``stuffed bear'' in the Teatime scene, a relevancy map can be rendered focusing on the location with maximum relevancy activation.}
     \label{fig:overview}
     \vspace{0.1cm}
 \end{center}
}]

\maketitle
\renewcommand{\thefootnote}{\fnsymbol{footnote}}
\footnotetext[2]{The corresponding author.}
\renewcommand{\thefootnote}{\arabic{footnote}}
\begin{abstract}
Recovering 3D structures with open-vocabulary scene understanding from 2D images is a fundamental but daunting task. Recent developments have achieved this by performing per-scene optimization with embedded language information. However, they heavily rely on the calibrated dense-view reconstruction paradigm, thereby suffering from severe rendering artifacts and implausible semantic synthesis when limited views are available. In this paper, we introduce a novel generative framework, coined \textbf{LangScene-X}, to unify and generate 3D consistent multi-modality information for reconstruction and understanding. Powered by the generative capability of creating more consistent novel observations, we can build generalizable 3D language-embedded scenes from only sparse views.
Specifically, we first train a TriMap video diffusion model that can generate appearance (RGBs), geometry (normals), and semantics (segmentation maps) from sparse inputs through progressive knowledge integration. Furthermore, we propose a Language Quantized Compressor (LQC), trained on large-scale image datasets, to efficiently encode language embeddings, enabling cross-scene generalization without per-scene retraining. Finally, we reconstruct the language surface fields by aligning language information onto the surface of 3D scenes, enabling open-ended language queries. Extensive experiments on real-world data demonstrate the superiority of our LangScene-X over state-of-the-art methods in terms of quality and generalizability. Project Page: \url{https://liuff19.github.io/LangScene-X/}.
\end{abstract}

    
\section{Introduction}
\label{sec:intro}
As spatial learners with strong prior knowledge, humans can perceive, interpret, and understand the 3D physical world from only few-shot visual captures. Therefore, it is fundamental and crucial in computer vision to learn 3D structures from images with scene understanding, which allows to interact and query the 3D worlds through open-ended language~\cite{azuma2022scanqa, gordon2018iqa, yang2024thinking, lerf}, offering a wide range of applications~\cite{zhang2024sam-e, rashid2023language-grasp, shan2023robustness-sam} such as robotics, autonomous navigation, and VR/AR.

With the rapid advancements in NeRF~\cite{nerf} and Gaussian Splatting~\cite{3dgs}), recent works~\cite{langsplat, lerf, li2024langsurf} have incorporated the language features from the CLIP~\cite{clip} into the 3D representations to build a 3D language field that supports open-vocabulary object querying. Although they can achieve promising results in per-scene optimization with calibrated dense views (usually more than 20 views) as input, they cannot generalize to unseen scenes and suffer from severe artifacts in 3D structures with implausible semantics when encountering insufficient input views~\cite{hu2024sparselgs}. 
Moreover, these approaches heavily rely on scene-specific auto-encoders to compress 512-dim CLIP features into lower-dim latent space for efficient rendering. Such reliance significantly increases training time and a tendency for domain overfit, which hinders fast inference and the ability to effectively scale with large datasets, limiting the range of applications in real-world scenarios.

The heart of high-quality language-embedded 3D scenes lies in the need to integrate multimodal information (\ie, semantics, geometry, and appearance) into a cohesive 3D representation. While dense, calibrated views provide abundant data for this integration, their acquisition is costly and impractical for broader adoption~\cite{hu2024sparselgs}. In contrast, constructing language-embedded fields from sparse views is inherently more challenging due to the scarcity of input information, yet it is critical for expanding applicability. The primary difficulty is extracting and fusing sufficient multimodal knowledge from limited inputs to achieve coherent 3D scene reconstruction and understanding.

To address this, we propose \textbf{LangScene-X}, a novel generative paradigm to build generalizable 3D language-embedded scenes from very sparse views (\ie, as few as two images). Building upon the representation learning capabilities of generative models~\cite{ye2024stablenormal, amit2021segdiff, ravishankar2024scaling-diff-perception}, our key insight is to unleash the strong generative prior to unify information for reconstruction and understanding in a single video diffusion model. Specifically, we first build a TriMap video diffusion model that can generate appearance (RGB images), geometry (normal maps), and semantics (semantic maps) from sparse inputs. To ensure 3D consistency across generated frames and to bridge domain gaps between different modalities (RGB, normals, and segmentation masks), we meticulously design a multi-task training strategy that progressively integrates the knowledge from these diverse domains. Powered by the strong generalizability of video diffusion, our TriMap video diffusion can generate hierarchical semantic maps at varying levels of granularity. 
To reduce the memory cost and enhance scalability for large-scale data, we propose a generalizable \underline{L}anguage \underline{Q}uantized \underline{C}ompressor (LQC) trained on large-scale datasets, which encodes high-dimensional language features as low-dimensional discrete indices without sacrificing essential properties. This approach eliminates the need for per-scene retraining, reduces memory overhead, and enables rapid rendering of language-embedded Gaussians.
Finally, we reconstruct the language-embedded surface fields by contextually aligning the generated semantics onto the surface (generated normals) of 3D scenes (generated RGBs), enabling a broad range of downstream tasks with 3D language. We summarize the contributions of the paper as follows:
\begin{itemize}
    \item We introduce LangScene-X, a novel generative framework that constructs generalizable 3D language-embedded fields from sparse views, which unify the information of scene reconstruction and understanding in a video diffusion model.
    \item We build the TriMap video diffusion model through a progressive multi-task training scheme, which can generate 3D-consistent RGB images, normal maps, and semantic maps across video frames. 
    \item {We propose a generalizable language quantized compressor (LQC) for efficient feature representation, reducing memory usage and rapid rendering while maintaining essential language features properties}.
    \item We conduct extensive experiments to verify the efficacy of our framework and show that LangScene-X outperforms existing methods for high quality and generalizability, revealing the great potential to craft 3D language fields from a generative paradigm.
\end{itemize}

\section{Related Work}
\label{sec:related work}

\subsection{Gaussian Splatting}
3D Gaussian Splatting (3DGS)~\cite{3dgs} has recently made significant strides in advancing 3D scene representation.
It offers high-resolution real-time rendering capabilities that surpass traditional Neural Radiance Field (NeRF) methods.
This efficiency facilitates various downstream applications~\cite{liu2024make, matsuki2024gaussian-slam, liu2024physics3d, lerf}. 
For scene surface reconstruction~\cite{sugar}, PGSR~\cite{pgsr} employs multi-view supervision to the detailed reconstruction of 3D surfaces and meshes.
In the field of scene understanding~\cite{scannet, liu2023semantic-ray}, LangSplat~\cite{langsplat} and LangSurf~\cite{li2024langsurf} introduce Language Embeddings into Gaussian attributes to facilitate efficient text querying in 3D space.
Despite its capabilities, 3DGS struggles with accurately reconstructing scenes with limited observed views~\cite{charatan2024pixelsplat}.
Several subsequent methods~\cite{li2024dngaussian,chen2024mvsplat} enhance this process by incorporating geometric regularization.
Moreover, other methods~\cite{lsm,gpnerf} utilize feed-forward models without per-scene optimization to perform generalized 3D 
Gaussian representations.
Recently, ViewCrafter~\cite{yu2024viewcrafter} and ReconX~\cite{liu2024reconx} leverage video diffusion models to interpolate nearby frames given single or two views to achieve dense view prediction and scene reconstruction.
In contrast, our method offers a multi-task generation pattern that generates appearance (RGBs), geometry (normals), and semantics (semantic maps) from sparse views, promoting efficient and accurate scene reconstruction and understanding.

\subsection{Video Diffusion Models}
Integrating additional constraints for conditional video generation has yielded remarkable results in the field of 3D scene reconstruction/generation. Much of the earlier work in video generation has concentrated on incorporating various control signals to video diffusion models. For instance, some works~\cite{he2024cameractrl,bahmani2024vd3d} introduce camera pose embedding into U-Net based video diffusion models~\cite{xing2024dynamicrafter} to drive controllable video generation. 
%
Moreover, ReconX~\cite{liu2024reconx} and ViewCrafter~\cite{yu2024viewcrafter} integrate DUSt3R~\cite{wang2024dust3r} as explicit 3D prior for 3D consistent video generation.
Additionally, DimensionX~\cite{sun2024dimensionx} achieves consistent video generation with large trajectory motion based on a powerful DiT-based video generation structure~\cite{yang2024cogvideox}. 
However, those works are limited to single modality generations, restricting their downstream applications. 
Rather than injecting additional control signals, some methods~\cite{hu2024depthcrafter,zhao2024motiondirector} customize video diffusion models by fine-tuning on a set of reference videos with similar patterns, such as depth maps or motions.
Yet, these methods often rely on complete video rather than sparse view inputs and involve complex network architectures and training procedures.
In contrast, we propose a TriMap video diffusion model, which can generate 3D consistent videos across various domains (\textit{i.e.}, appearance, geometry, and semantics) by progressively fusing the cross-domain knowledge into the DiT-based video diffusion model~\cite{yang2024cogvideox} without sacrificing performance.

\subsection{Language Embedded Fields}
The basic idea of the language-embedded Gaussian representation is to inject 2D language features from CLIP~\cite{clip} and SAM~\cite{sam} into Gaussians and then fascinating real-time text-guided object query in novel views. 
However, directly encoding 512-dim CLIP features into Gaussian primitives is memory expensive.
LangSplat~\cite{langsplat} adopts Autoencoder to encode scene-wise CLIP features into low-dimension latent for Gaussian training.
%
OpenGaussian~\cite{wu2024opengaussian} utilizes a coarse-to-fine feature codebook to register CLIP features into a low-dimension index.
Nevertheless, such methods heavily rely on scene-specific feature compressors, which is time-consuming and hinders extension or generalization to other scenes.
In contrast, our method proposes a Quantized Language Compressor for compact feature representation, which maps high-dimension features into a low-dim index while maintaining essential feature properties. 
Moreover, they require calibrated dense views to reconstruct the scene, leading to poor performance in the scenario of sparse or limited view input.
Some generalizable methods~\cite{gpnerf,lsm} directly encode multi-image inputs into 3D semantic representation through large foundation models~\cite{attention} for novel view generation. 
GP-NeRF~\cite{gpnerf} adopts Dual Transformer to aggregate RGB and semantic features from input views into an implicit representation.
LSM~\cite{lsm} integrates learning-based MASt3R~\cite{wang2024dust3r} for dense language Gaussian predictions from sparse unposed images.
However, these methods are limited to category-specific domains, such as objects and indoor scenes, and are prone to artifacts due to their limited model representation capabilities.
Our method unleashes the power of video diffusion models in a novel way to generate 3D consistent RGB, appearance prediction, and segmentation masks simultaneously, offering scalability and generalization with improved performance.

\section{Method}
\label{sec: method}
\subsection{Overview of LangScene-X}
\begin{figure*}
    \centering
    \includegraphics[width=1\linewidth]{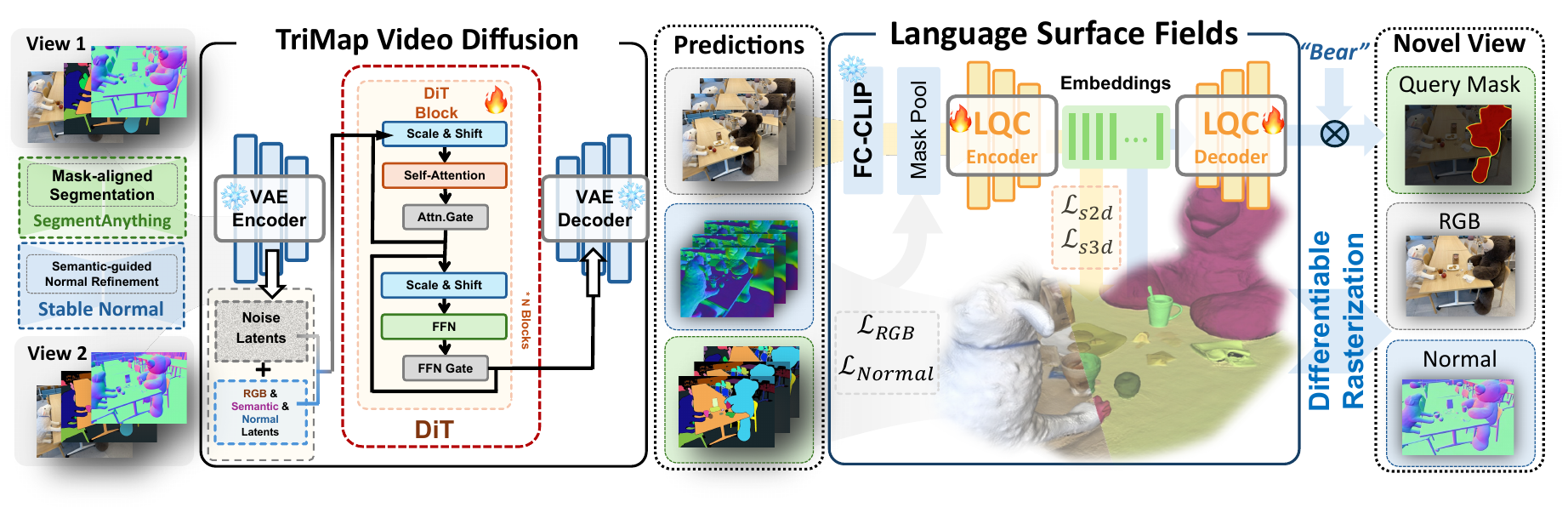}
    \caption{\textbf{Pipeline of LangScene-X.} Given two sparse-view images as input, we first generate a sequence of 3D consistent RGB images, normal maps, and segmentation maps from TriMap video diffusion model, which provides the dense frames for later 3D scene reconstruction and understanding. Then we project high-dimensional semantic features into low-dimensional discrete space through a generalizable Language Quantized Compressor (LQC). Finally, we reconstruct the 3D language-embedded scenes with generated and compressed information, which supports open-ended language queries in any viewpoint.}
    \label{fig:framework-pipeline}
\end{figure*}

Given $N$ sparse views (\ie, as few as two images) as input, our goal is to reconstruct and understand the underlying 3D scene (\ie, construct the language-embedded surface fields). In our framework LangScene-X, we first build the TriMap video diffusion model to generate 3D consistent RGB images, normal maps, and semantic maps from sparse-view input (Sec.~\ref{subsec: 3D-context video}), which provides the dense frames for reconstruction and understanding. Then we compress the high-dimensional language features into low-dimensional discrete space through a generalizable language quantized compressor (Sec.~\ref{subsec: 3D-context video}). This eliminates per-scene retraining and enables rapid rendering of Gaussians. Finally, we reconstruct the language-embedded 3D scenes with generated and compressed information (Sec.~\ref{subsec: language-embedded recon}), supporting open-ended language queries in any viewpoint. Our pipeline is depicted in Figure~\ref{fig:framework-pipeline}.
\begin{figure}
    \centering
    \includegraphics[width=\linewidth]{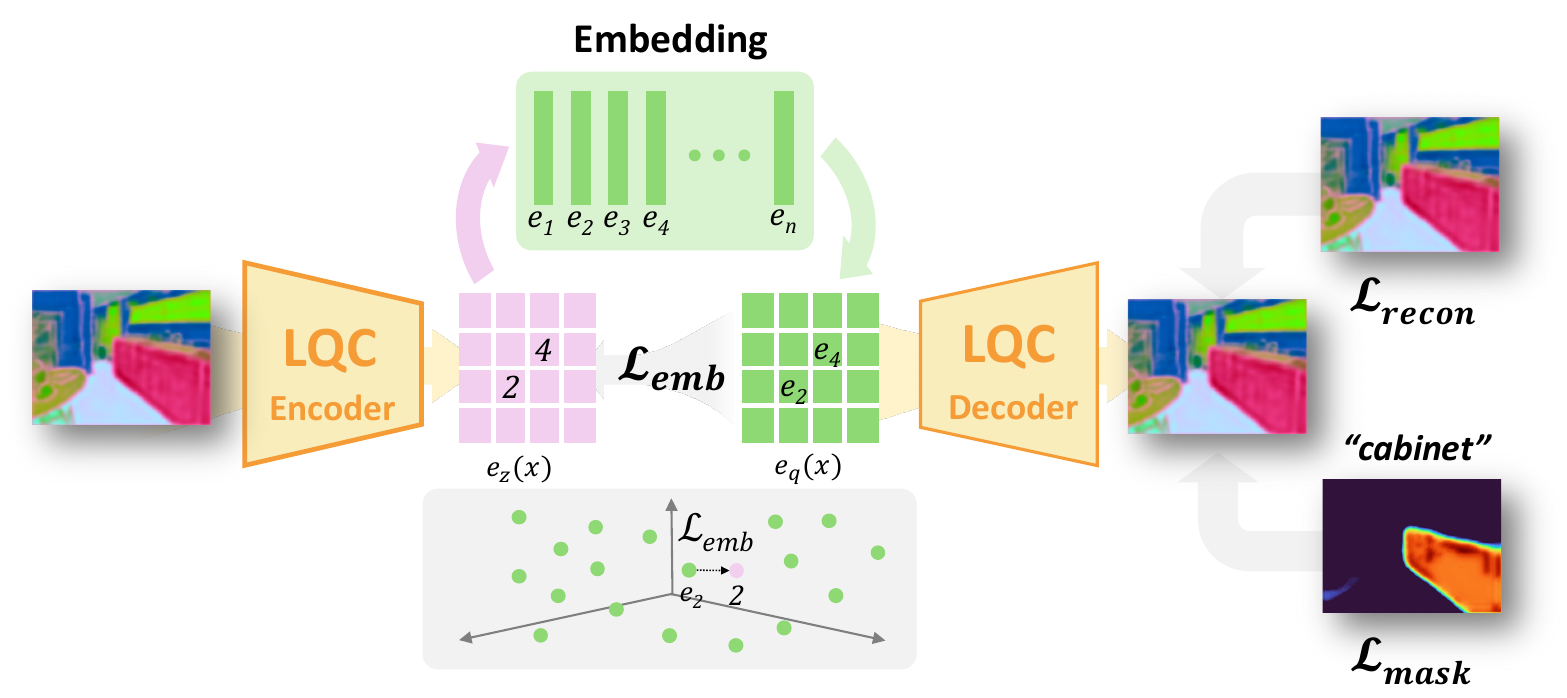}
    \caption{The illustration of Language Quantized Compressor (LQC). By leveraging learnable embedding and vector quantisation strategy, it compresses high-dimensional language features into discrete \(D\)-channel latent representation, facilitating efficient language fields reconstructions and render.}
    \label{fig:lqc}
\end{figure}
\subsection{Building the TriMap Video Diffusion}
\label{subsec: 3D-context video}
Existing works like LERF~\cite{lerf} and LangSplat~\cite{langsplat} essentially require very dense multi-view inputs with accurate calibration due to inherent per-scene optimization schemes from NeRF~\cite{nerf} or 3D Gaussian~\cite{3dgs}. As a result, these methods often struggle to synthesize high-quality images with semantics from insufficient views, especially in unseen areas. Recently, video generative models have shown promise for generating frames featuring 3D structure~\cite{voleti2024sv3d, liu2024reconx, yu2024viewcrafter}
and perception capability~\cite{shao2024learning-video-depth, ravishankar2024scaling-diff-perception}. This inspires us to unleash the strong generative prior of large pre-trained video diffusion models to generate frames that unify the information for reconstructing (RGB images and normal maps) and understanding (segmentation masks) 3D scenes in language-embedded fields. 
However, it is non-trivial to achieve 3D consistency in video generation across different perception patterns~\cite{liu2024reconx, shao2024learning-video-depth}.

To address this, we build the large TriMap video diffusion model to generate RGB images, normals, and semantic maps from only sparse views (\ie, as few as two images). We follow the architecture as CogVideoX~\cite{yang2024cogvideox}, which is a transformer-based video diffusion model~\cite{peebles2023scalable-dit}. Before we train the TriMap video diffusion, we first modify the i2v architecture to support key-frame interpolation, which is more flexible for sparse-view input. Given a $T$-frame video $V\in \mathbb{R}^{H \times W \times 3}$, we choose the first $I_f$ and the last image $I_e$ padded with zeros to obtain a condition video with the same size as $V$. Then, we encode the condition video with a causal VAE~\cite{yang2024cogvideox} encoder to get a latent vector concatenated with a Gaussian noise of the same size. Finally, we iteratively denoise the noise latent and apply the VAE decoder to obtain key-frame interpolation results. Keeping the architecture as unchanged as possible, we can fully leverage the pretrained prior knowledge.
Let $\mathcal{D}_{I}$, $\mathcal{D}_{N}$ and $\mathcal{D}_{S}$ be different domain mappers $\mathcal{M}$ for RGB images, normals and semantic maps respectively (\ie, $\mathcal{M}:\mathbb{R}^{H \times W \times C} \times \mathcal{D}_{\theta} \rightarrow \mathbb{R}^{H\times W \times C}$) while keeping the same channels, we meticulously divide the training into four stages. (1) We first use large-scale web data to train the TriMap video diffusion to support a general key-frame interpolation task. (2) We then finetune the model with a small amount of 3D consistent video data ($\sim 10K$) to learn 3D consistency across frames. (3) Next, we annotate 200 clips of 3D consistent video data with $\mathcal{D}_N$ to get normal videos along with RGB videos to finetune the model. (4) Finally, we apply $\mathcal{D}_S$ to annotate 300 clips of semantic masks from 3D consistent video data and finetune with RGB and normal videos. The overall training objective is:
\begin{equation}
    \mathcal{L}_{\text{diff}} = \mathbb{E}_{\boldsymbol{x},\epsilon \sim \mathcal{N}(0, I), t, \mathcal{D}_{i}\in \mathcal{M}}\left[\left\|\epsilon-\epsilon_\theta\left(\boldsymbol{x}_t, t, \mathcal{D}_{i}\right)\right\|_2^2\right] ,
\end{equation}
where $\boldsymbol{x}_t$ is the noise latent from the ground-truth views of the training data. Through empirical studies, such progressive knowledge integration training strategy can facilitate 3D consistency and multi-task perception in generated frames simultaneously. Moreover, powered by strong generalizability, our TriMap video diffusion can generate multi-hierarchy masks $\{ \mathbb{M}^h| h=s, m, l \}$ at inference time from only two input views segmented by $\mathcal{D}_{S}$, where $s, m, l$ represents small, medium, and large hierarchy levels of the segmentation masks.

\begin{table*}[t]
\centering
\caption{\textbf{2D Quantitative Results on LERF-OVS Dataset.} We report the open-vocabulary localization accuracy (\%) and 2D semantic segmentation (IoU scores). LSeg~\cite{lseg} is a 2D open-vocabulary segmentation network, while other methods~\cite{langsplat,li2024langsurf} are per-scene optimized language field models. LSM~\cite{lsm} is the generalizable language Gaussian method. The \textbf{bold} denotes the best results.}
\label{tab:lerf2dseg}
\resizebox{1\textwidth}{!}{
\begin{tabular}{ c | c c | c c | c c | c c | c c }
\toprule
\multirow{2}{*}{Scene Type} & \multicolumn{2}{c|}{LSeg~\cite{lseg}} & \multicolumn{2}{c|}{LangSplat~\cite{langsplat}} & \multicolumn{2}{c|}{LangSurf~\cite{li2024langsurf}} & \multicolumn{2}{c|}{LSM~\cite{lsm}}  & \multicolumn{2}{c}{LangScene-X (Ours)} \\
 & mAcc\(\uparrow\) & mIou\(\uparrow\) & mAcc\(\uparrow\) & mIou\(\uparrow\) & mAcc\(\uparrow\) & mIou\(\uparrow\) & mAcc\(\uparrow\) & mIou\(\uparrow\) & mAcc\(\uparrow\) & mIou\(\uparrow\) \\
\midrule
Teatime & 65.22& 30.58& 30.43& 15.81& 35.57& 18.82& 44.46& 19.62& \textbf{78.91} & \textbf{45.07} \\
Ramen  & 54.55& 37.86& 49.45& 15.08& 48.85& 21.79& 54.55& 23.01& \textbf{72.73}& \textbf{42.92} \\
Kitchen& 72.73& 51.37& 27.27& 15.23& 36.36& 18.72& 49.99& 26.05& \textbf{90.91} & \textbf{63.58} \\ 
\midrule
Overall& 64.17& 39.94& 35.72& 15.37& 40.26& 19.78& 49.67& 22.89& \textbf{80.85}& \textbf{50.52} \\
\bottomrule
\end{tabular}}
\end{table*}  

\begin{table*}[t]
\centering
\caption{\textbf{2D Quantitative Results on ScanNet Dataset.} We report the open-vocabulary localization accuracy (\%) and 2D semantic segmentation (IoU scores). The \textbf{bold} denotes the best results.}
\label{tab:scannet2dseg}
\resizebox{1\textwidth}{!}{
\begin{tabular}{ c | c c |c c |c c |c c |c c }
\toprule
\multirow{2}{*}{Scene Type} & \multicolumn{2}{c|}{LSeg~\cite{lseg}} & \multicolumn{2}{c|}{LangSplat~\cite{langsplat}} & \multicolumn{2}{c|}{LangSurf~\cite{li2024langsurf}} & \multicolumn{2}{c|}{LSM~\cite{lsm}}  & \multicolumn{2}{c}{LangScene-X (Ours)} \\
 & mAcc\(\uparrow\) & mIou\(\uparrow\) & mAcc\(\uparrow\) & mIou\(\uparrow\) & mAcc\(\uparrow\) & mIou\(\uparrow\) & mAcc\(\uparrow\) & mIou\(\uparrow\) & mAcc\(\uparrow\) & mIou\(\uparrow\) \\
\midrule
0085\_00   & 63.64& 28.71& 42.73& 28.98& 68.18& 30.47& 67.65& 39.09& \textbf{95.45}& \textbf{51.68} \\
0114\_02   & 83.32& 65.97& 40.21& 15.75& 78.08& 63.59& 73.33& 43.42& \textbf{92.35}& \textbf{72.10} \\
0616\_00   & 84.62& 66.96& 46.15& 11.35& 68.34& 44.30& 78.65& 56.41& \textbf{97.85}& \textbf{70.71} \\
0617\_00   & 86.36& 62.66& 45.45& 29.09& 72.73& 56.04& 75.63& 54.33& \textbf{90.91}& \textbf{71.65} \\ \midrule
Overall & 79.49& 56.08& 43.64& 21.29& 71.83& 48.60& 73.82& 48.31& \textbf{94.14}& \textbf{66.54} \\
\bottomrule
\end{tabular}
}
\end{table*}

\subsection{Language Quantized Compressor}
\label{subsec: LQC}
Previous methods~\cite{langsplat, li2024langsurf} rely on scene-specific autoencoders to compress high-dimensional language features into low-dimensional features with per-scene optimization. 
Such a technique represents latent with continuous distribution, which struggles to represent essential language properties with extremely low-dimension compression (\textit{i.e.}, 3-channels) on large-scale data. 
Meanwhile, language features are inherently discrete~\cite{van2017neural} (\textit{i.e.}, features with same categories share consistent distributions).  
Therefore, discrete representations are naturally suited for compressing language features during the process of low-dimensional representation.
To this end, as shown in Figure.~\ref{fig:lqc}, we leverage Vector Quantization into our compressor and propose a Language-Quantized Compressor (LQC) trained on large-scale datasets (COCO~\cite{lin2014coco}).
Specifically, we define a learnable embeddings \(E = \{e_1, e_2, \cdots, e_k|e_k \in \mathbb{R}^{D}\}\) to capture essential language properties during the medium layer of our compressor, where \(k\) denotes the size of discrete latent space and \(D\) denotes the channel of the latents. 
The language features \(x\) produced by off-the-shelf dense CLIP feature extractor~\cite{lseg} pass through the encoder to obtain \(z_e(x)\),  which serves as indices to calculate nearest neighbor look-up with embeddings \(E\) and map \(z_e(x)\) into $1\sim k$ embeddings:
\begin{equation}
z_q(x) = e_k, \, \text{where} \, k = \operatorname{argmin}_j \left\| z_e(x) - e_j \right\|_2,
\end{equation}
where \(z_q(x)\) is the final mapped discrete latent. 
After that, \(z_q(x)\) passes through the decoder and obtains the reconstructed feature \(\hat{x}\).
However, training such a compressor is non-trivial, as the look-up operation obstructs the gradient flow from the decoder to the encoder.
To address it, we directly copy the gradient flow from decoder to encoder networks for encoder-decoder training, where :
\begin{equation}
\mathcal{L}_{\textit{r}} = \left\| x - \operatorname{decoder} \left( z_e(x) + \operatorname{sg}\left( z_q(x) - z_e(x) \right) \right) \right\|_2^2,
\end{equation}
where sg means the operation of ``stop gradient''.  For learnable embeddings training, we utilize classic dictionary learning algorithms that push embeddings \(E\) towards encoder outputs \(z_e(x)\):
\begin{equation}
\mathcal{L}_{emb} = \left\|\operatorname{sg}\left[z_e(x)\right]-E\right\|_2^2.
\end{equation}
Additionally, to ensure accurate language-feature alignment, we utilize pseudo-mask supervision by applying L2 loss on text-guided activation maps between vanilla feature \(x\) and reconstructed feature \(\hat{x}\):
\begin{equation}
\mathcal{L}_{mask} = \left\|\hat{x}\cdot T - x\cdot T\right\|_2^2
\end{equation}
The overall criterion can be summarized as follows:
\begin{equation}
    \mathcal{L}_{lqc} = \lambda_1\mathcal{L}_{r} + \lambda_2\mathcal{L}_{emb} + \lambda_3\mathcal{L}_{mask},
\end{equation}
where \(\lambda_1, \lambda_2, \lambda_3\) denote loss weights.
Such strategies enable high-ratio feature compression for subsequent language Gaussian training, preserving all essential language properties.

\subsection{Language-Embeded Surface Fields}
\label{subsec: language-embedded recon}
Building upon the 3D-consistent frames generation (RGB, normals, and semantic maps) and efficient compress pipeline, we are able to construct accurate language-embedded surface fields that facilitate efficient novel-view synthesis and open-ended 3D language-guided query.
Specifically, given the generated RGB sequences \(\mathbf{C}\in \mathbb{R}^{D\times H\times W\times 3}\), we initialize the sparse point clouds of the scene using DUSt3R~\cite{wang2024dust3r} and train our language surface fields using regular L2 RGB loss \(\mathcal{L}_{rgb}\) with several steps. 
To ensure accurate 3D language representation, we then perform joint training using geometry and semantic supervision to construct language-embedded surface fields.
In pratice, we leverage the powerful normal priors \(\mathbf{N}\in \mathbb{R}^{D\times H\times W\times 3}\) generated by the TriMap video diffusion, we adopt a progressive normal regularization  \(\mathcal{L}_{normal}\) to optimize the geometry representation of our model:
\begin{equation}
\mathcal{L}_{normal}= \begin{cases}\left\|\mathbf{N}_p-\mathbf{N}\right\|_1 & ,\quad \text{step} <T_n \\ \left\|\mathbf{N}_p-\hat{\mathbf{N}}\right\|_1\end{cases},
\end{equation}
where \(\mathbf{N}_p\) is the rendered normal by our training fields and \(T_n\) is the hyper-parameter step. \(\hat{\mathbf{N}}\) is the generated normal that filters out uncertain regions that hard to optimize ( \(\theta_n > \tau_{thr}\) ) to alleviate the impact of incorrect normal generation. The \(\tau_{thr}\) is the hyper-parameter threshold and \(\theta_n\) is the angle difference between \(\mathbf{N}_p\) and \(\mathbf{N}\):
\begin{equation}
\theta_p=\arccos \left(\frac{\mathbf{N}_p \cdot \mathbf{N}}{\left\|\mathbf{N}_p\right\|\left\|\mathbf{N}\right\|}\right).
\end{equation}
For semantic supervision, we utilize pre-trained dense CLIP extractor~\cite{yu2023convolutions} to extract language features from generated images \(\mathbf{C}\) and pass through our proposed LQC to obtain 3-channel language latent. 
Moreover, to facilitate accurate 3D semantic representation, apart from regular L2 semantic loss, we additionally adopt a 2D and 3D clustering criterion based on the generated segmentation masks \(\mathbf{M}\):
\begin{equation}
    \begin{cases}
        \mathcal{L}_{s2d} = \|\hat{\mathbf{F}}_{lang}(v_1) - \hat{\mathbf{F}}_{lang}(v_2)\|_2,  \; v_1, v_2 \in \mathbf{M} \\
        \mathcal{L}_{s3d} = \sum_{i=1}^{N} \sum_{k=1}^{K} \mathbf{f}^{sem}_k \log \left( \mathbf{f}^{sem}_k / \mathbf{f}^{sem}_j \right), 
    \end{cases}
\end{equation}
where \(\hat{\mathbf{F}}^{lang}\) is the rendered feature map and \(\mathbf{f}^{sem}\) is the feature attribute on the Gaussian. Such a strategy facilitates the language Gaussians are closely attached on the surface of corresponding objects.
\section{Experiment}
\label{sec: experiment}

\begin{figure*}[t]
    \centering
    \includegraphics[width=0.9\linewidth]{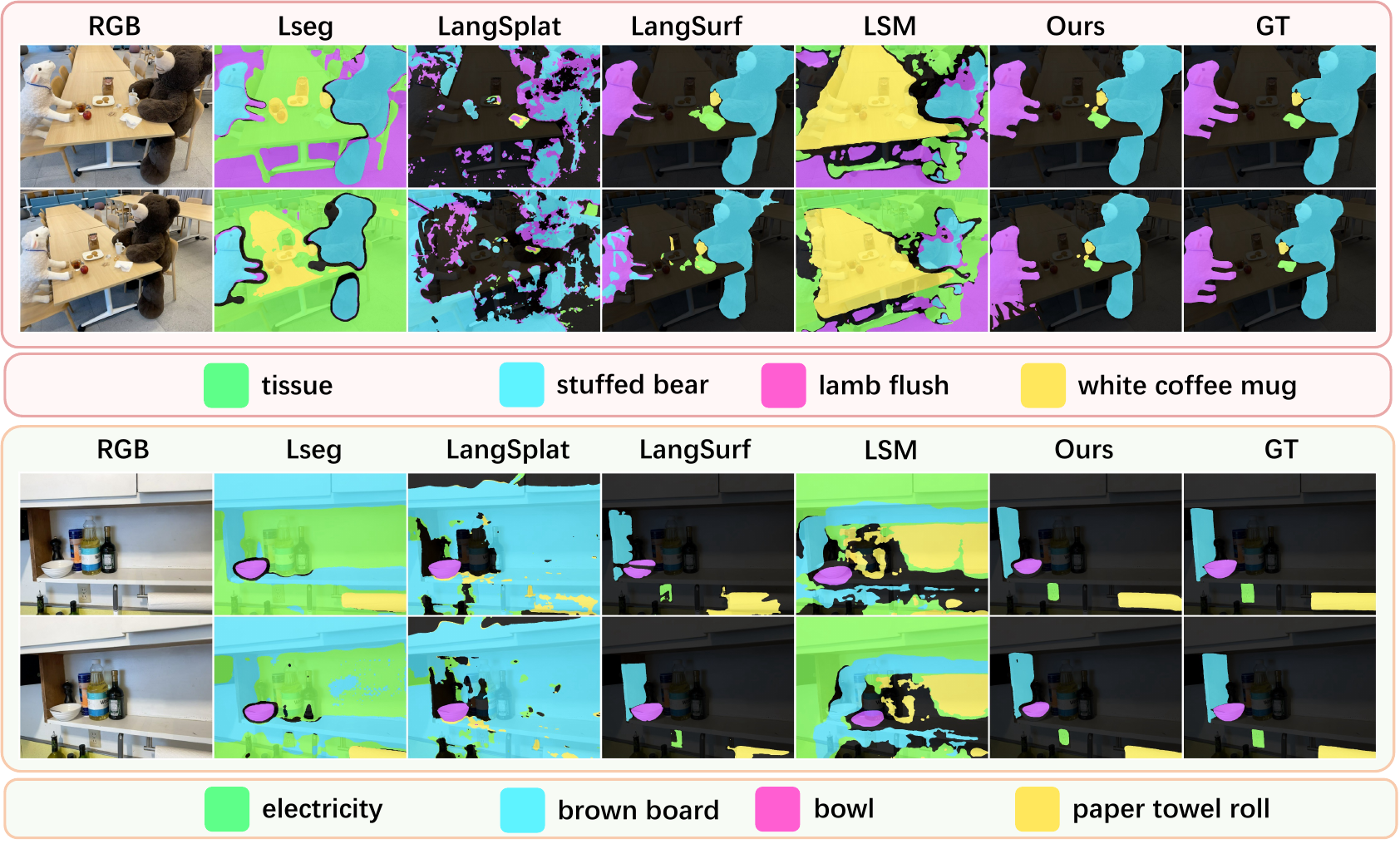}
    \vspace{-0.1in}
    \caption{\textbf{2D Segmentation Results on LERF-OVS~\cite{lerf} Dataset.} Here, we showcase two cases (\textit{i.e.}, Teatime, Kitchen) with multiple segmentation masks with text query. On the top, we display the rendered results of our method and other methods, along with the corresponding ground truth annotations. On the bottom,  we present the images alongside the queried texts.}
    \label{fig:langscene_ovs}
    \vspace{-0.1in}
\end{figure*}

\begin{figure*}[htbp]
    \centering
    \includegraphics[width=0.9\linewidth]{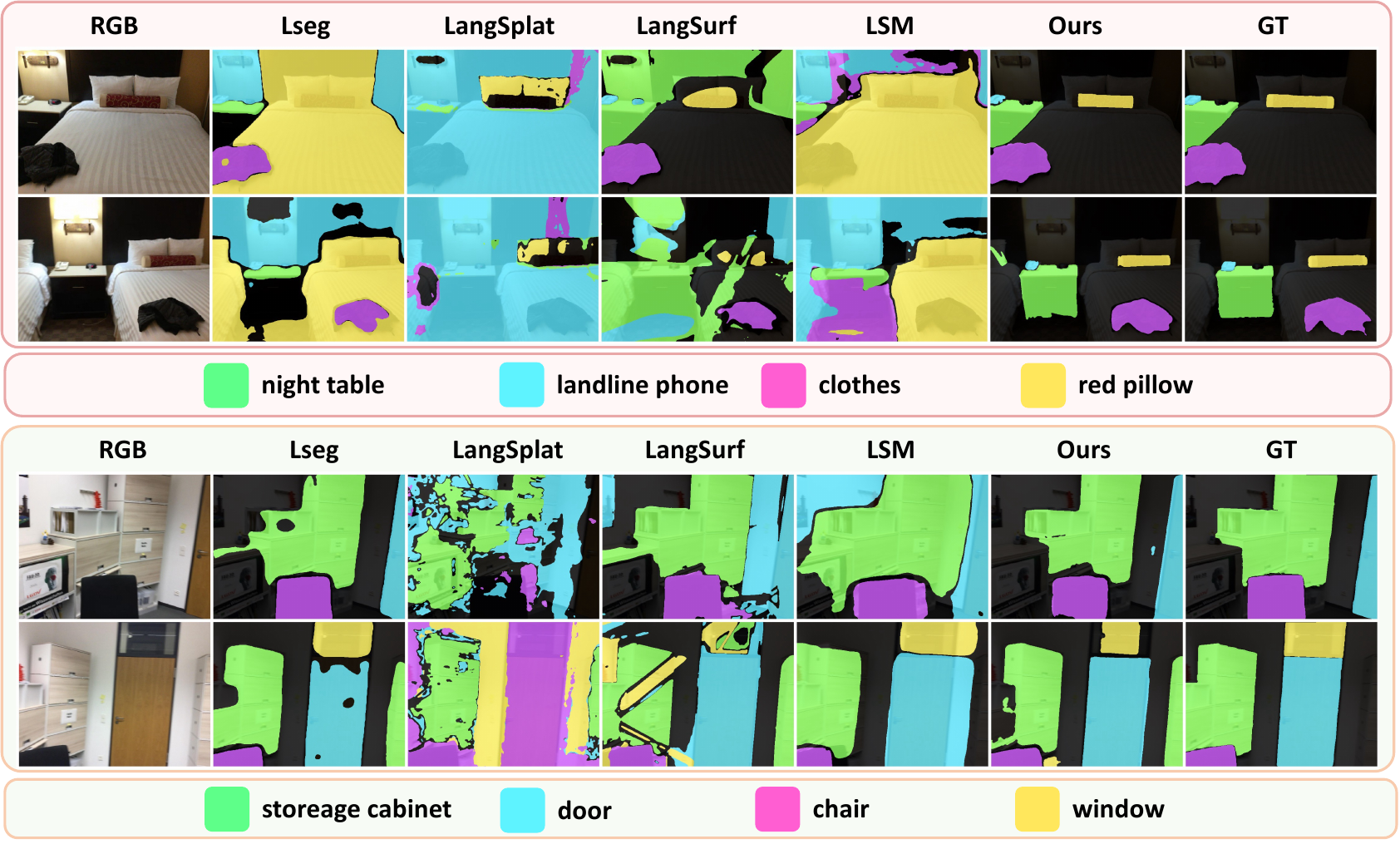}
    \vspace{-0.1in}
    \caption{\textbf{2D Segmentation Results on Scannet~\cite{scannet} Dataset.} Here, we showcase two cases (\textit{i.e.}, 0085\_00, 0114\_00) with multiple segmentation masks with text query. The masks predicted by ours contain more comprehensive regions and sharper boundaries than other methods, such as the ``Cabinet'' prompt, which also surpasses the GT masks.}
    \label{fig:langscene_scannet}
\end{figure*}

\subsection{Experiment Setup}
\noindent \textbf{Implementation Details.} We conduct all experiments on  8 NVIDIA A800 (80G) GPUs. In the framework of LangScene-X, we choose CogVideoX~\cite{yang2024cogvideox} as the backbone architecture of our TriMap video diffusion, StableNormal~\cite{ye2024stablenormal} as our normal mapper $\mathcal{D}_N$, and SAM2~\cite{ravi2024sam-2} as the semantic mapper $\mathcal{D}_S$. We first train TriMap video diffusion with the key-frame interpolation capability for one week on large-scale web data. Then we finetune it on 3D-consistent real data (\ie, RealEstate-10K~\cite{zhou2018stereo-re10k} and ACID~\cite{liu2021infinite-ACID}) with 2000 steps on the learning rate $1\times 10^{-5}$. Next, we apply StableNormal to annotate 200 normal video clips of 3D scene data from RealEstate-10k and finetune TriMap video diffusion along with RGB videos with 800 steps. Finally, we apply SAM2 to annotate 300 clips of semantic video clips to fine-tune with RGB and normal videos with 1000 steps. The AdamW~\cite{loshchilov2017decoupled-adamw} optimizer is employed for optimization. All videos are center-cropped and resized to $720\times 480$ resolution with $49$ frames. Additionally, for the Language Quantized Compressor, we set the number of embeddings as \(K=2048\) and the channel as \(D=3\). Then, we train our model on large-scale open-world dataset COCO~\cite{lin2014coco} with a batch size of 16 and 500,000 steps. We set loss weights \(\lambda_1 = 1, \lambda_2 = 0.2, \lambda_3 = 0.5\) during the training. For Language Surface Fields training, we first train the Gaussian model with only RGB and normal loss \(\mathcal{L}_{normal}\) for 5,000 steps, and then we utilize semantic losses (\textbf{e.g.}, \(\mathcal{L}_{s2d}, \mathcal{L}_{s3d}\) for 5,000 steps.

\begin{figure}[htbp]
    \centering
    \includegraphics[width=1\linewidth]{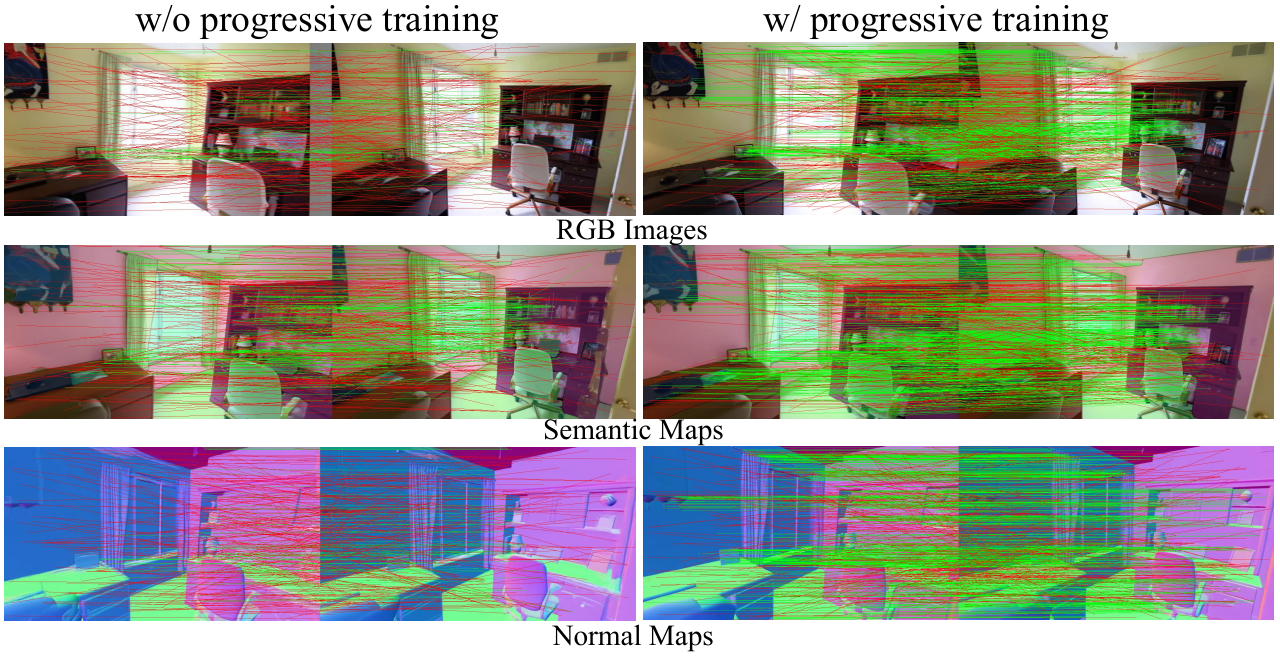}
    \vspace{-0.2in}
    \caption{\textbf{Feature Matching} comparison between our method and vanilla video diffusion mdoel .}
    \label{fig:feature-matching}
    \vspace{-0.2in}
\end{figure}

\noindent \textbf{Baseline and Metrics.} To demonstrate our strong capability in building 3D language-embedded scenes from only sparse views, we compare our LangScene-X against four competitive baselines: LSeg~\cite{lseg}, LangSplat~\cite{langsplat}, LangSurf~\cite{li2024langsurf}, and LSM~\cite{lsm}. For fair comparison, we follow the dataset choice in LangSplat~\cite{langsplat} and LangSurf~\cite{li2024langsurf} and conduct comparisons on LERF-OVS dataset~\cite{lerf} and ScanNet dataset~\cite{scannet}. The LERF dataset is an in-the-wild dataset captured by a handheld device, while ScanNet is a large scene dataset captured by RGB-D devices in complex indoor scenes. Each scene contains semantic labels at 3D level, making it suitable for 3D scene reconstruction and understanding tasks. For quantitative results, we report the standard metrics in semantic understanding, including open-vocabulary localization accuracy (mAcc) and semantic segmentation (mIoU scores).

\subsection{Main Results}
We conduct experiments on 2D open-vocabulary segmentation by querying the language features of each Gasussian with text on both LERF-OVS~\cite{kerr2023lerf} and Scannet~\cite{scannet} datasets, as reported in Tab.~\ref{tab:lerf2dseg} and Tab.~\ref{tab:scannet2dseg}. By comparing with existing state-of-the-art 3D language field techniques (\textit{e.g.}, LangSplat, LangSurf), unified 3D representation method (\textit{i.e.}, LSM), and 
open-vocabulary methods like LSeg, our method achieves superior performance in segmentation accuracy in both mIoU and mAcc metrics with a large margin, \textit{i.e.}, a \(10.58\%\) in terms of mIoU and a \(31.18\%\) in terms of mAcc on LERF-OVS dataset. On the ScanNet dataset, the improvement upon the best existing method comes to 14.92\% in terms of mIoU. The visualization of the 2D segmentation masks is shown in Fig.~\ref{fig:langscene_ovs} and Fig.~\ref{fig:langscene_scannet}. As can be seen, our approach excels both optimized-based and feed-forward-based methods by segmenting objects with fine-grained boundaries. This demonstrates the capability of our method for embedding more generated multi-modality priors with 3D consistency into the 3D space.

\begin{figure}[t]
    \centering
    \includegraphics[width=1\linewidth]{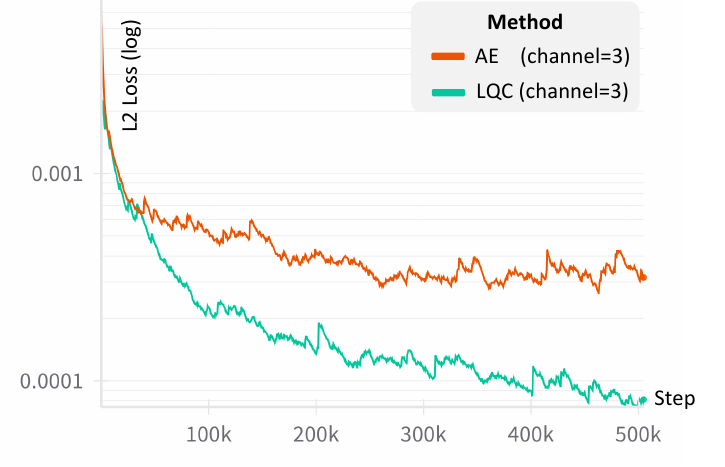}
    \vspace{-0.1in}
    \caption{\textbf{Training Curve} comparison between our LQC and regular autoencoder technique.}
    \label{fig:lqc_curve}
\end{figure}

\begin{figure}[htbp]
    \centering
    \includegraphics[width=1\linewidth]{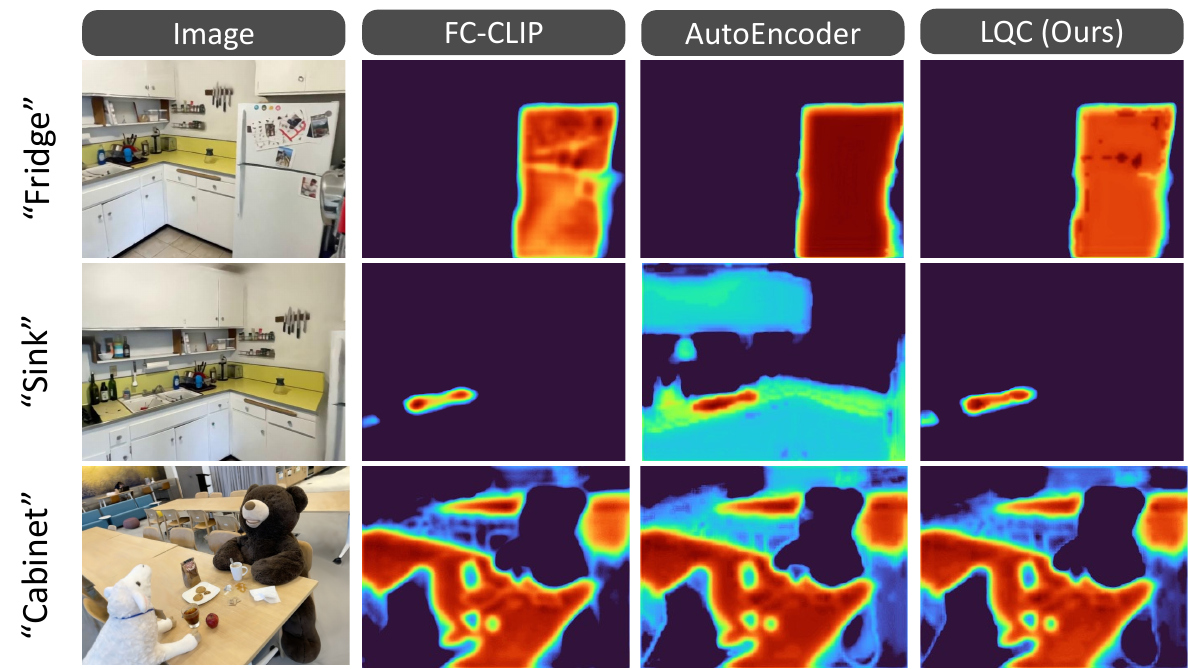}
    \caption{\textbf{Qualitative Comparison on LERF-OVS~\cite{lerf}}. We visualize text-query activation masks with various text prompts.}
    \label{fig:lqc_image}
\end{figure}

\subsection{Ablations}
We conduct ablation experiments with our TriMap Video Diffusion and Language Quantized Compressor techniques.

\noindent \textbf{TriMap Video Diffusion.} To further assess the geometric consistency of generated frames from TriMap video diffusion, we evaluate camera geometry alignment between frames. Specifically, we extract two frames at regular intervals from each video, creating pairs of two-view images. For each pair, we use a matching algorithm~\cite{ng2003sift} to find corresponding points and use RANSAC~\cite{fischler1981ransac} to filter out incorrect matches. Figure~\ref{fig:feature-matching} shows the effectiveness of progressive training in TriMap video diffusion, which achieves more matched points.


\noindent \textbf{Language Quantized Compressor.} We visualize the training curve of our method and traditional autoencoder. As shown in Fig.~\ref{fig:lqc_curve}, it is evident that our quantized technique demonstrates superior performance in both loss convergence rate and accuracy in terms of L2 loss (from \(1e^{-3}\) to \(1e^{-4}\)). Moreover, we showcase the text-query activation masks in Fig.~\ref{fig:lqc_image}, where our method enable to perform sharper boundaries and more accurate activation scores within the query objects. This indicates the effectiveness of our discrete latent representation of LQC to preserve essential language properties during compression.

\begin{table}[t]
\footnotesize
    \caption{\textbf{Ablations of proposed module and losses.} We perform ablations on both ScanNet (scene0085)~\cite{scannet} and LERF (Teaime)~\cite{lerf} with the segmentation metric mIoU.}
    \centering
\begin{tabular}{cccc|c|c}
\toprule
Progressive Train & LQC & \(\mathcal{L}_{s2d}\) & \(\mathcal{L}_{s3d}\) & ScanNet & LERF  \\ \midrule
\XSolidBrush          & \Checkmark                             & \Checkmark                            & \Checkmark                             & 44.25& 39.04\\
\Checkmark            & \XSolidBrush                           & \Checkmark                            & \Checkmark                             & 44.59& 41.56\\
\Checkmark            & \Checkmark                           & \XSolidBrush                            & \Checkmark                             & 40.05& 36.26\\
\Checkmark            & \Checkmark                             & \Checkmark                            & \Checkmark                             & \textbf{51.68}& \textbf{45.07}\\ \bottomrule
\end{tabular}
\label{tab:ablation}
\vspace{-5mm}
\end{table}

\section{Conclusion}
In this paper, we present LangScene-X, a generative framework that builds generalizable 3D language-embedded fields from only sparse views, which unify the information of reconstructing and understanding scenes in one video diffusion model. Specifically, we first train a TriMap video diffusion model through progressive knowledge integration, which can generate 3D consistent RGBs, normals, and semantic maps. Then we introduce a language quantized compressor to map high-dimensional language features into efficient feature representations. Finally, we reconstruct the language-embedded Gaussians by aligning the generated semantics onto the surface of 3D scenes. We believe LangScene-X provides a promising direction for 3D reconstruction and understanding through a generative paradigm.


{
    \small
    \bibliographystyle{ieeenat_fullname}
    \bibliography{main}
}


\end{document}